\documentclass[conference]{IEEEtran}
\IEEEoverridecommandlockouts
\usepackage{cite}
\usepackage{amsmath,amssymb,amsfonts}
\usepackage{algorithmic}
\usepackage{graphicx}
\usepackage{textcomp}
\usepackage{xcolor}
\def\BibTeX{{\rm B\kern-.05em{\sc i\kern-.025em b}\kern-.08em
    T\kern-.1667em\lower.7ex\hbox{E}\kern-.125emX}}
\begin{document}

\title{An Advanced Framework for Ultra-Realistic Simulation and Digital Twinning for Autonomous Vehicles\\}

\author{\IEEEauthorblockN{1\textsuperscript{st} Yuankai He}
\IEEEauthorblockA{\textit{Computer Science} \\
\textit{University of Delaware}\\
Newark, USA \\
willhe@udel.edu}
\and
\IEEEauthorblockN{2\textsuperscript{nd} Hanlin Chen}
\IEEEauthorblockA{\textit{Buildings and Transportation Science Division} \\
\textit{Oak Ridge National Laboratory}\\
Oak Ridge, USA \\
chenh1@ornl.gov}
\and
\IEEEauthorblockN{3\textsuperscript{rd} Weisong Shi}
\IEEEauthorblockA{\textit{Computer Science} \\
\textit{University of Delaware}\\
Newark, USA \\
weisong@udel.edu}
}

\maketitle

\begin{abstract}
Simulation is a fundamental tool in developing autonomous vehicles, enabling rigorous testing without the logistical and safety challenges associated with real-world trials. As autonomous vehicle technologies evolve and public safety demands increase, advanced, realistic simulation frameworks are critical. Current testing paradigms employ a mix of general-purpose and specialized simulators, such as CARLA and IVRESS, to achieve high-fidelity results. However, these tools often struggle with compatibility due to differing platform, hardware, and software requirements, severely hampering their combined effectiveness. This paper introduces BlueICE, an advanced framework for ultra-realistic simulation and digital twinning, to address these challenges. BlueICE's innovative architecture allows for the decoupling of computing platforms, hardware, and software dependencies while offering researchers customizable testing environments to meet diverse fidelity needs. Key features include containerization to ensure compatibility across different systems, a unified communication bridge for seamless integration of various simulation tools, and synchronized orchestration of input and output across simulators. This framework facilitates the development of sophisticated digital twins for autonomous vehicle testing and sets a new standard in simulation accuracy and flexibility. The paper further explores the application of BlueICE in two distinct case studies: the ICAT indoor testbed and the STAR campus outdoor testbed at the University of Delaware. These case studies demonstrate BlueICE’s capability to create sophisticated digital twins for autonomous vehicle testing and underline its potential as a standardized testbed for future autonomous driving technologies.
\end{abstract}


\section{Introduction}

With the evolving landscape of modern transportation, the role of Connected and Autonomous Vehicles (CAVs) has become increasingly significant. The widespread deployment of CAV systems offers substantial benefits in terms of mobility, safety, and energy efficiency \cite{fhwaCAV}. However, poorly designed CAV systems can lead to catastrophic failures, diminishing public confidence in this promising technology \cite{noy2018automated}. Thus, rigorous validation and testing are essential before CAVs can be fully integrated into the transportation network \cite{zhao2016accelerated,9831031,feng2023dense}.
Since field testing in open environments is both costly and risky, simulation emerges as a crucial component in this process \cite{8500545}. It enables researchers to evaluate the performance of autonomous driving algorithms and vehicle systems under controlled conditions, with digital twins further ensure the soundness of simulation result by utilizing real-world data to replicate complex virtual scenarios \cite{veledar2019digital}.

Game engine-based simulators like Grand Theft Auto V \cite{RefWorks:RefID:16-martinez2017gta} and CARLA \cite{RefWorks:RefID:6-dosovitskiy2017carla:} have been pivotal in early autonomous vehicle testing; and recently, the Department of Transportation (DOT) has developed simulators such as CDASim\cite{RefWorks:RefID:22-2024cdasim:} and CARMA to provide more focused integration of traffic, network, and vehicle dynamics. However, these simulators lack the necessary physical accuracy for crucial environmental interactions like tire tracks, snow, and water dynamics, which are critical for realistic simulation scenarios.


In response to this need, this paper introduces Blue\textbf{ICE}—an innovative \textbf{Integrative Customizable} simulation \textbf{Environment} that significantly advances beyond existing tools. BlueICE offers the ability to \textit{integrate and synchonize} a wide spectrum of simulation capabilities. It allows for the seamless integration of environments like CARLA with NS-3 and SUMO, and specialized tools such as IVRESS, which are essential for simulating specific environmental effects like off-road tire tracks and water dynamics.

BlueICE’s architecture effectively decouples dependencies between different computing platforms, hardware, and software versions, addressing major integration challenges. Through containerization, it isolates each simulator's operational environment, preventing conflicts and enhancing operational stability across diverse systems. A unified communication bridge within BlueICE facilitates smooth interactions among different simulators, ensuring data consistency and synchronization. This not only simplifies the management of complex simulations but also significantly reduces the integration effort and time, making it an efficient tool for researchers.

The following sections will detail the architecture of BlueICE and discuss specific case studies demonstrating its application in creating advanced digital twins for autonomous vehicle testing. We will explore how BlueICE’s innovative features contribute to establishing a robust, scalable, and customizable simulation environment, setting a new standard in the field and paving the way for future developments in autonomous vehicle research and testing.

 The rest of the paper is organized as follows: Section \ref{sec:related_works} discusses some of the most popular related works and their major shortcomings. Section \ref{sec:system_overview} gives an overview of the BlueICE architecture and elaborates on each component. Section \ref{sec:case_study} demonstrates our state-of-the-art digital twin. Section \ref{sec:conclusion} concludes the paper and discusses our vision.
\section{Related Works}\label{sec:related_works}


\subsection{Summary of simulators}
Simulators have been foundational to the research and development of connected and autonomous vehicles (CAVs). The original definition of simulator in 1956 is "a computer model to represent real systems" \cite{goode1956use}. They allow for the analysis of interactions between the vehicle under test (VUT) and other traffic participants, addressing complex "what if" scenarios that are either too risky or costly to test in real life \cite{kaur2021survey}.

As the simulation landscape evolved, traditional platforms like Grand Theft Auto V (GTA V) and those based on Unreal Engine and Unity provided foundational support for early testing. However, these simulators often fall short in accurately modeling environmental interactions such as tire tracks, snow, and water dynamics—elements critical for assessing vehicle responses under various conditions.

\begin{table*}[!t]
  \caption{Comparing Different Simulators}
  \label{tab:comparison}
  \centering
  \begin{tabular}{@{}r||c|c|c|c|c|c|c|c|c|c @{}} 
    \hline
    \textbf{Simulator} & \textbf{Veh} & \textbf{DynAGT} & \textbf{Sens} & \textbf{Env} & \textbf{Traffic} & \textbf{Sce}& \textbf{Co-Sim} & \textbf{OpenSrc} & \textbf{Cloud} & \textbf{Network} \\ 
    \hline    
    \tt{GTA V\cite{RefWorks:RefID:16-martinez2017gta}} & \textasciitilde & x  & x & \textasciitilde & x & x & x & x & x & x \\
    \hline
    \tt{Simulink\cite{RefWorks:RefID:18-xue2022modeling}} & \checkmark & x  & \textasciitilde & \textasciitilde & \checkmark & \checkmark & \textasciitilde & x & x & x \\
    \hline
    \tt{PreScan\cite{RefWorks:RefID:19-leneman2013prescan}} & \checkmark & x  & \textasciitilde & \textasciitilde & \checkmark & \checkmark & \textasciitilde & x & x & x \\
    \hline
    \tt{TORCS\cite{RefWorks:RefID:17-espie2019torcs}} & \checkmark & x  & \textasciitilde & \textasciitilde & x & \checkmark & x & \checkmark & x & x \\
    \hline
    \tt{Apollo\cite{RefWorks:RefID:21-2023apollo:}} & \checkmark & \checkmark  & \textasciitilde & \textasciitilde & \checkmark & \checkmark & \textasciitilde & \checkmark & x & x \\
    \hline
    \tt{LG SVL\cite{RefWorks:RefID:10-rong2020lgsvl}} & \checkmark & \checkmark  & \textasciitilde & \textasciitilde & \checkmark & \checkmark & \textasciitilde & \checkmark & x & x \\
    \hline
    \tt{CARLA\cite{RefWorks:RefID:6-dosovitskiy2017carla:}} & \checkmark & \checkmark  & \textasciitilde & \textasciitilde & \checkmark & \checkmark & \textasciitilde & \checkmark & x & x \\
    \hline
    \tt{Gazebo\cite{RefWorks:RefID:22-koenig2004design}} & \checkmark & \checkmark  & \textasciitilde & \textasciitilde & \checkmark & \checkmark & \textasciitilde & \checkmark & x & x \\
    \hline
    \tt{CarSim\cite{carsim}} & \checkmark & \checkmark  & \textasciitilde & \textasciitilde & \checkmark & \checkmark & \textasciitilde & \checkmark & x & x \\
    \hline
    \tt{MOSAIC\cite{RefWorks:RefID:15-schunemann2011v2x}} & \checkmark & \checkmark  & \textasciitilde & \textasciitilde & \checkmark & \checkmark & \textasciitilde & \checkmark & x & \checkmark\\
    \hline
    \tt{CARMA\cite{RefWorks:RefID:9-lochrane2020carma:}} & \checkmark & \checkmark  & \textasciitilde & \textasciitilde & \checkmark & \checkmark & \textasciitilde & \checkmark & x  & \checkmark\\
    \hline
    \tt{AWSim\cite{RefWorks:RefID:12-2022tier}} & \checkmark & \checkmark  & \textasciitilde & \textasciitilde & \checkmark & \checkmark & x & x & x & x\\
    \hline
    \tt{DriveSim\cite{RefWorks:RefID:24-nvidia2023nvidia}} & \checkmark & \checkmark  & \checkmark & \textasciitilde & \checkmark & \checkmark & \textasciitilde & \checkmark & x & x\\
    \hline
    \tt{VISTA\cite{RefWorks:RefID:25-amini2022vista}} & x & \checkmark  & \checkmark & \textasciitilde & \checkmark & \checkmark & x & \checkmark & x & x\\
    \hline
    \tt{IVRESS\cite{RefWorks:RefID:3-ivress}} & \checkmark & x  & \checkmark & \checkmark & x & x & x & x & x & x\\
    \hline
    \tt{DDT\cite{RefWorks:RefID:4-dynamic}} & \checkmark & x  & \checkmark & \checkmark & x & x & x & x & x & x\\
    \hline
    \tt{\textbf{\textit{BlueICE (Ours)}}} & \checkmark & \checkmark  & \checkmark & \checkmark & \checkmark & \checkmark & \checkmark & \checkmark & \checkmark & \checkmark \\
    \hline
  \end{tabular}
\end{table*}

This gap has spurred the development of high-fidelity simulations that can more precisely model-specific vehicle behaviors. Table \ref{tab:comparison} compares popular simulators across essential testing categories: vehicle physics, dynamic agents, sensor data, environment simulation, traffic simulation, scenario customization, network, and co-simulation. The table marks each simulator with symbols indicating full support (\checkmark), no support (x), and limited support (\textasciitilde).

Vehicle physics, crucial for simulating autonomous vehicles, varies significantly across different models; a truck, for example, will generate more air resistance than a sedan. Most simulators offer vehicle customization to ensure accurate replication of these physical behaviors. Additionally, simulating dynamic agents and traffic flows is vital for validating AV responses in realistic traffic conditions. Although dedicated traffic simulators are proficient at these tasks, some vehicle simulators also handle them effectively with a higher computational cost.

Sensor modeling is arguably the most vital aspect of autonomous vehicle simulation. Sensor data quality greatly affects the reliability of algorithms developed in simulated environments. Unfortunately, many simulators only provide medium-fidelity sensor data, which may not suffice for detailed testing requirements, leading to research questions involving sim-to-real domain knowledge transfer.

Environmental simulation is another critical area where many simulators struggle. While they can often replicate various weather and lighting conditions in certain kinds of sensor data, they are incapable of doing the same task for other kinds of sensors. Also, accurately modeling complex environmental details like tire tracks and water dynamics remains challenging. This underscores the importance of specialized simulators that can handle these high-level details to ensure vehicle safety under diverse conditions.

Microscopic traffic simulators, such as VISSIM, AIMSUN, Paramics, and SUMO, provide detailed control over groups of individual vehicle behaviors and are indispensable for creating realistic traffic scenarios. They enhance the richness of traffic environments, enabling CAVs to navigate more realistically and response to background vehicles whose maneuvers are bounded by traffic flow models.

In contrast, vehicle-level simulators such as LGSVL, CARLA, and Gazebo focus primarily on the ego vehicle’s performance and simulation. While they excel in this area, their ability to control background traffic may not be as comprehensive as that microscopic traffic simulators offer.

Integrating these varied tools into a unified testing framework is challenging. BlueICE addresses this need with an architecture designed to facilitate seamless co-simulation for multi-modal simulators, enhancing both testing and development of CAVs. The next sections will detail BlueICE's architecture and discuss its role in advancing simulation standards for autonomous vehicle testing.
\section{BlueICE Architecture}\label{sec:system_overview}

BlueICE is an advanced simulation framework designed to facilitate the integration of diverse simulators into a unified system for testing and validating autonomous vehicles. By enabling the coordination of multiple simulations running on different operating systems and hardware, BlueICE addresses the challenges associated with complex, multi-simulator environments.

\begin{figure}[htbp]
\centering
\includegraphics[width=1.0\columnwidth]{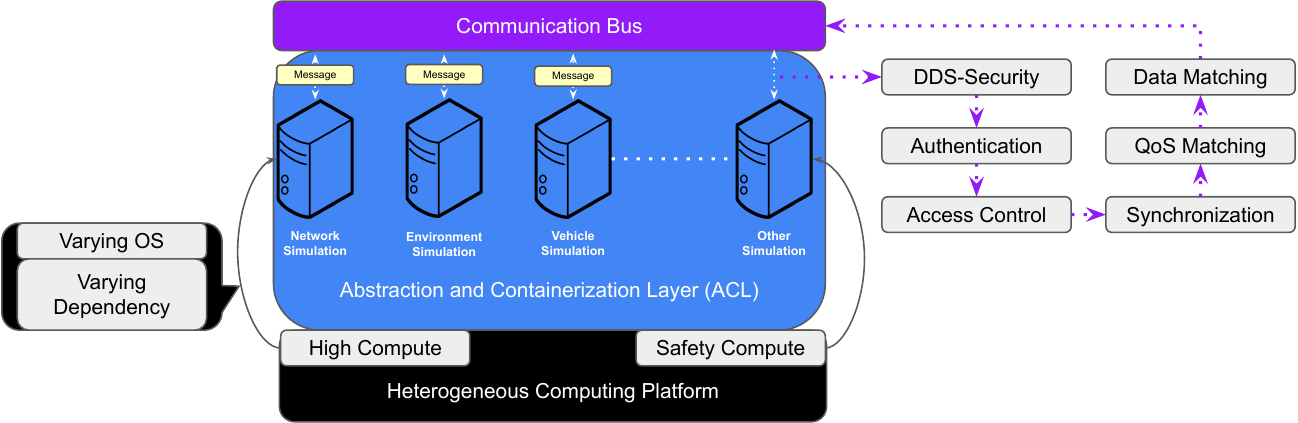}
\caption{Framework Layer Overview. The heterogeneous computing layer lays the groundwork to support different operating systems and different dependencies such as software version and system architecture; the containerization layers house different modular simulators; the communication layer offers a bus for security, authentication, access control, data matching, and data synchronization.}
\label{fig: overview}
\end{figure}

\subsubsection{Framework Layers}

Fig \ref{fig: overview} illustrates the layered architecture of the BlueICE framework, depicting how various simulators are encapsulated within individual containers and interconnected through the communication layer, highlighting the modular and scalable design of the system.
\begin{itemize}
    \item \textbf{Computing Layer} - BlueICE leverages a heterogeneous network of computers to run various simulators. This setup accommodates the unique computational needs of each simulator, allowing them to perform optimally on hardware and operating systems best suited to their requirements.
    \item \textbf{Containerization Layer} - To ensure that each simulator functions independently and does not interfere with others, BlueICE employs containerization. This strategy isolates simulators from each other, making system updates and integrating new simulators smoother and less disruptive to ongoing operations.
    \item \textbf{Communication Layer} - The integration and synchronization across these modular simulators are achieved through a robust communication framework provided by ROS (Robot Operating System). ROS enables effective time synchronization and message exchange across the network, ensuring that all simulators can operate in synchronization despite being on separate systems.
\end{itemize}

\subsubsection{Modular Simulation Environments}
BlueICE's core strength lies in its modular approach to simulator integration. Each simulator is treated as a modular component that can be independently managed and operated.

\subsubsection{Security and Data Integrity}
BlueICE ensures robust security and data integrity by leveraging the ROS security framework and DDS-SECURITY specifications, which are crucial for protecting sensitive data and maintaining the integrity of operations within its distributed architecture. Secure communication is done with ROS DDS-SECURITY to encrypt communications between nodes, protecting data from eavesdropping and tampering. Authentication and authorization mechanisms ensure that only verified nodes can access the network, safeguarding against unauthorized interventions.

Access control is strictly enforced through detailed policies defined by DDS-SECURITY, which restrict operations and interactions among nodes, enhancing overall system security. The ROS logging capabilities support effective monitoring, aiding in the prompt detection and response to potential security issues.

\subsubsection{Scalability and Flexibility}

BlueICE can easily accommodate additional simulators or expand computational resources to meet growing testing requirements. Its modular design allows for the easy integration of new technologies and methodologies, ensuring the framework remains relevant as simulation technology evolves.

\section{Case Study}\label{sec:case_study}

\subsection{Digital Twin at the University of Delaware D-STAR Campus}

\begin{figure}[htbp]
\centering
\includegraphics[width=1.0\columnwidth]{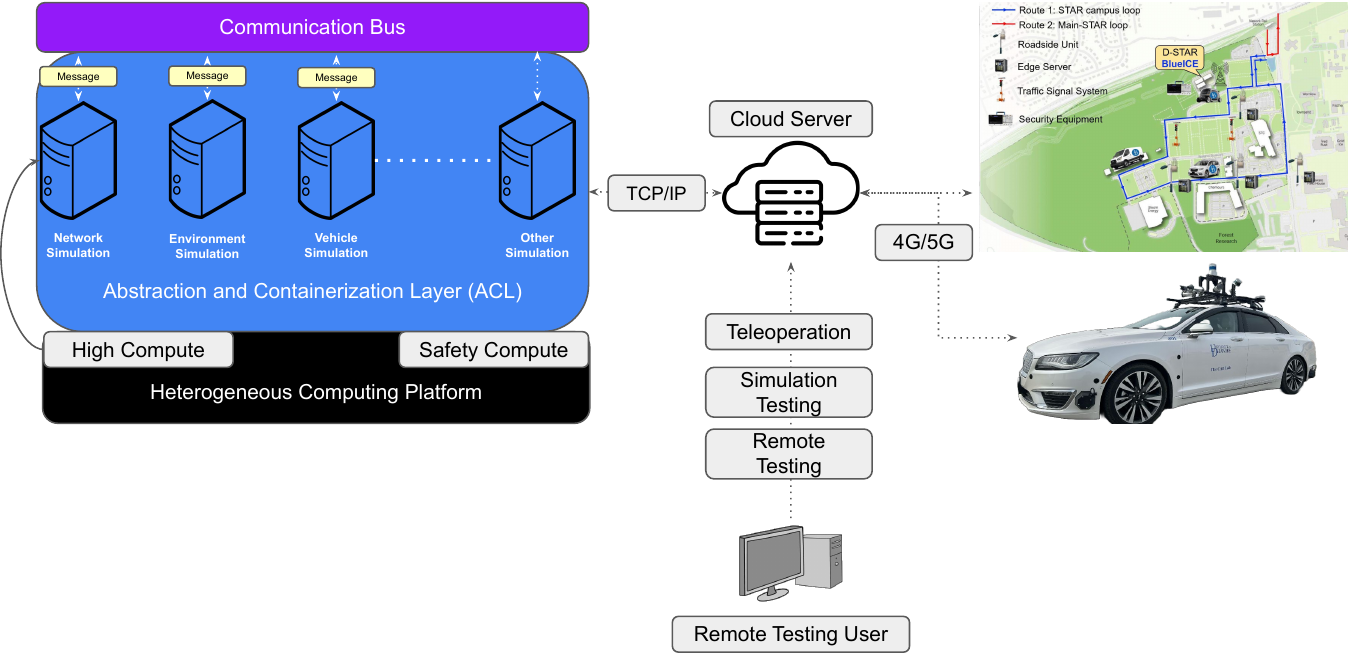}
\caption{D-STAR DT and BlueICE Integration Overview. The simulation server is hosted on a cloud service, with each simulator occupying a different container environment. Real-world data such as network latency, vehicle position and results, and other relevant information are sent through our private 5G network. The remote user can either test their algorithms in simulation or remotely control their vehicle on the ground, given a safety driver is present. }
\label{fig: uddt}
\end{figure}

This section presents our digital twin's preliminary designs and findings at the University of Delaware's D-STAR campus. Equipped with a Cisco private 5G network, C-V2X solutions from iSmartways and Danlaw Inc., and DELL edge servers, the D-STAR campus is a campus-scale mobility testbed that supports extensive research and educational initiatives in connected and sustainable mobility. Fig \ref{fig: uddt} showcases our current and future design to improve the D-STAR testbed.

\subsubsection{Digital Twin Integration with BlueICE}
The digital twin utilizes the BlueICE framework to integrate real-world data to simulate autonomous driving scenarios. This system manages a variety of simulators across different hardware, including GPU-intensive simulators for vehicle dynamics and CPU-intensive simulators for network and traffic management. This integration is crucial for accurately replicating the complex environment of the D-STAR campus.

\subsubsection{Dynamic Data Synchronization and Environmental Modeling}

Environmental elements such as roads and buildings are initially modeled using OpenStreetMap, Roadrunner, and Google Satellite images and are \textit{continuously} updated based on sensor data from LiDAR and cameras. This dynamic updating ensures the digital twin accurately mirrors the real-world conditions of the campus, adjusting to changes as they occur.

\subsubsection{Evaluation and Results}

\begin{figure}[htbp]
\centering
\includegraphics[width=1.4\columnwidth]{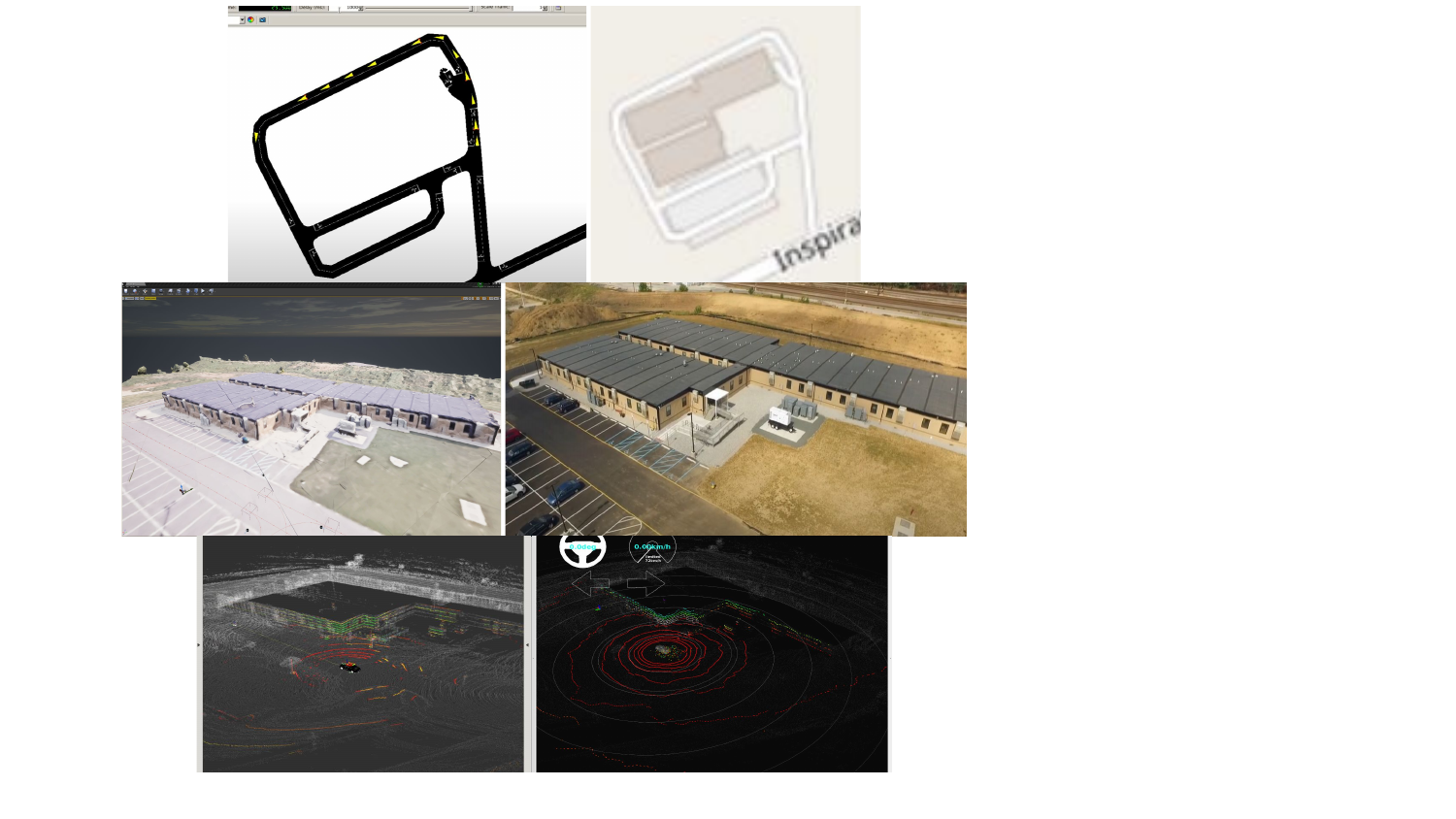}
\caption{Comparision Between Simulated (Left) and Real (Right). From top to bottom: road network, buildings environment, and sensor data}
\label{fig: udcomp}
\end{figure}

Fig \ref{fig: udcomp} illustrates the comparison between the simulated world and the real world. The efficacy of the digital twin is demonstrated through its ability to use simulated point clouds for vehicle localization in real-world environments. This critical evaluation involves comparing the vehicle’s actual positions, given by GPS and LiDAR localization, against its positions determined through simulated data. \textit{The mean-square-root error between the real and simulated positions is less than 5 cm.}

This precise localization is crucial, ensuring that the simulations can reliably mimic real-world vehicle conditions. This allows researchers to conduct thorough and meaningful tests without the risks associated with physical trials. The success of this evaluation not only proves the digital twin's functionality but also showcases its potential as a powerful tool for advancing autonomous vehicle technologies.

\subsubsection{Future Enhancements and Research Directions}

Looking ahead, the D-STAR digital twin will incorporate more detailed pedestrian movements and complex traffic scenarios to enhance its testing capabilities. These developments will enable more comprehensive evaluations of autonomous vehicle systems under varied and dynamic conditions, further supporting the advancement of safe and efficient autonomous transportation technologies.

\subsubsection{Conclusion}

The D-STAR digital twin, powered by the BlueICE framework, exemplifies the potential of integrated, real-world data simulations in developing and testing autonomous vehicles. By continuously evolving to reflect actual campus changes and proving capable in critical tests like real-world localization, the digital twin provides a vital platform for pushing the boundaries of connected and autonomous mobility research.

\subsection{Indoor Connected and Autonomous Testbed (ICAT) Digital Twin at the University of Delaware}

\begin{figure}[htbp]
\centering
\includegraphics[width=1.0\columnwidth]{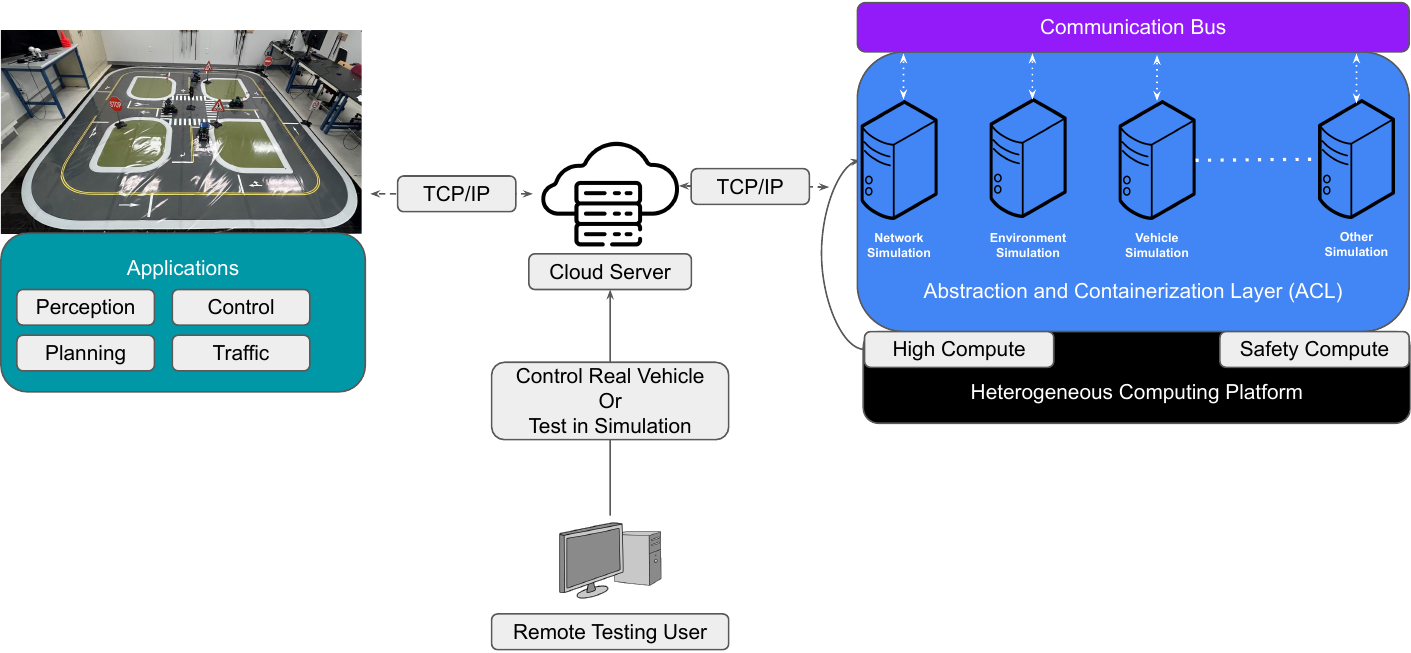}
\caption{ICAT DT and BlueICE Integration Overview. Applications are run on the infrastructure overseeing the ICAT testbed. The recognized results are uploaded to the simulation environment running on a cloud server. Remote testing can be conducted to control test vehicles directly or experiment in simulation.}
\label{fig: icatoverview}
\end{figure}

This section explores the ICAT testbed\cite{RefWorks:RefID:28-icat2024}, our indoor facility designed for the detailed study and validation of connected and autonomous vehicle technologies under controlled conditions. Unlike the dynamic and expansive setup of the D-STAR campus, ICAT focuses on a more contained environment where maps and vehicle models remain constant, allowing for precise experiments and analyses. Fig \ref{fig: icatoverview} illustrates how BlueICE connects the physical world, simulated world, and remote testers together.

\subsubsection{Integration of Digital Twin with Constant Elements}

In ICAT, the digital twin operates with a set of constant elements, including predefined maps and vehicle models. This stability is crucial for conducting repeatable experiments where specific variables can be isolated and tested. The maps within the digital twin include detailed renderings of the indoor test environment, complete with necessary traffic signage and infrastructure layouts.

\subsubsection{Dynamic Interaction and Map Updating}
One of the unique features of the ICAT digital twin is its dynamic interaction system, where an overseeing infrastructure is responsible for monitoring and updating the map. This infrastructure detects changes, such as the presence or absence of stop signs, and updates the map accordingly. This capability ensures that any temporary changes to the test environment are immediately reflected in the digital twin, maintaining the accuracy of the simulation.

\subsubsection{Vehicle and Infrastructure Communication}
The test vehicle plays a critical role within this environment. It processes localization information based on real-time interactions within the testbed and sends this data to the infrastructure. This information is used for global path planning and vehicle control, integrating the vehicle's operations with the overall management of the testbed. This setup tests the vehicle's ability to navigate and respond to the environment accurately and its capability to communicate effectively with the controlling infrastructure.

\subsubsection{Evaluation and Results}
The effectiveness of the ICAT digital twin is evaluated based on its ability to simulate and respond accurately to indoor testing conditions. Key metrics include the precision of map updates, vehicle localization reliability, and communication effectiveness between the vehicle and the infrastructure. These factors ensure the digital twin can reliably replicate and extend the test results to real-world scenarios. \textit{With BlueICE, ICAT is able to keep the vehicle localization mean-square-root error below 2 cm, map update latency to under 1 second, and communication delay between real and cloud below 11 ms with a CDF of 0.91666.} A cumulative distribution of communication latency is shown in Fig \ref{fig: icat_commu_cdf}:

\begin{figure}[htbp]
\centering
\includegraphics[width=1.0\columnwidth]{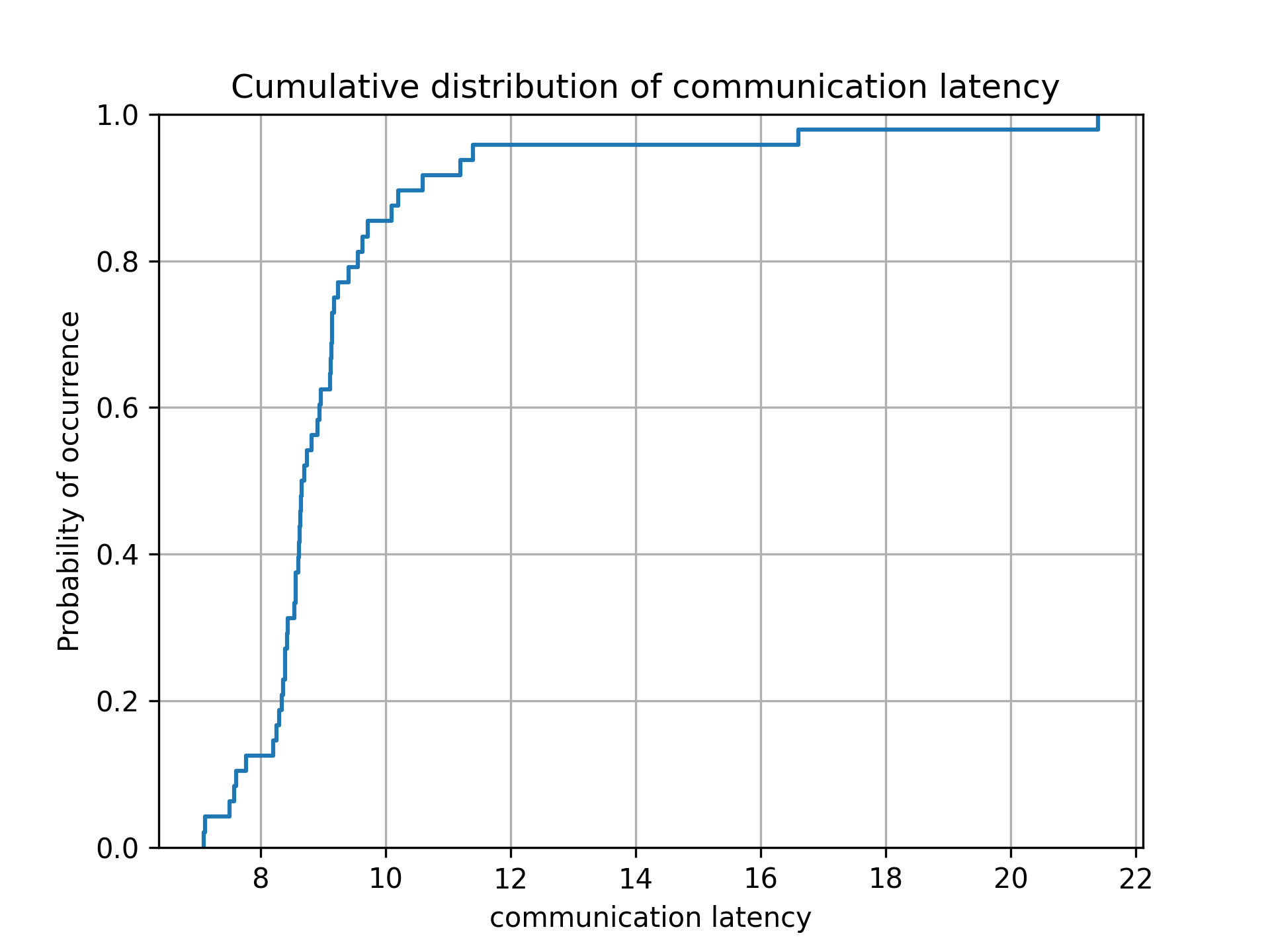}
\caption{ICAT Physical Testbed Communication Latency to Simulation Server Hosted on the Cloud}
\label{fig: icat_commu_cdf}
\end{figure}

\subsubsection{Future Enhancements and Research Directions}
ICAT aims to incorporate more complex interactions and introduce variable elements into the testbed, such as dynamic obstacles or altering traffic patterns. These enhancements will challenge the digital twin's adaptability and improve its capability to handle unexpected or changing conditions.

\subsubsection{Conclusion}
The ICAT indoor connected and autonomous testbed represents a vital component of our research into autonomous vehicle technologies. By maintaining certain constant elements while allowing for controlled updates and interactions, ICAT provides a robust platform for in-depth testing and development. The insights gained from this indoor environment are invaluable for advancing our understanding of autonomous vehicle behavior in a controlled yet adaptable setting.
\section{Conclusion}\label{sec:conclusion}
In this paper, we proposed the BlueICE framework. By leveraging containerization and a unified communication bridge, BlueICE facilitates seamless, synchronized communication across diverse simulation environments. This modular and scalable framework not only supports the integration and modification of various simulation tools but also ensures high fidelity and real-time performance across different hardware and software configurations. The case study conducted at the University of Delaware's D-STAR campus further exemplifies BlueICE's capabilities in creating a comprehensive digital twin environment that bridges the gap between simulation and real-world testing. As the landscape of transportation technology continues to evolve, BlueICE stands out as a robust solution for researchers seeking to innovate and test new transportation solutions with greater flexibility.

\section{Acknowledgement}
This work is partially supported by US National Science Foundation NSF-2311087.

\bibliographystyle{IEEEtran}
\bibliography{main}
\vspace{12pt}

\end{document}